\documentclass{article}

\usepackage{arxiv}

\usepackage[utf8]{inputenc}
\usepackage[T1]{fontenc}
\usepackage{hyperref}
\usepackage{url}
\usepackage{graphicx}
\usepackage{natbib}
\usepackage{doi}
\usepackage{booktabs}
\usepackage{amsmath}
\usepackage{amsfonts}
\usepackage{amssymb}
\usepackage{nicefrac}
\usepackage{microtype}
\usepackage{tabularx}
\usepackage{array}
\usepackage{makecell}
\usepackage{multirow}
\usepackage{float}
\usepackage{placeins}
\usepackage{flafter}

\newcolumntype{Y}{>{\centering\arraybackslash}X}
\title{UniPolymer: A Unified Framework for Property Prediction, Structure Recommendation, and Evaluation in Polyimide Design}

\author{
	Junquan Hu \\
	School of Software \\
	Dalian University of Technology \\
	Dalian, China \\
	\texttt{hjunquan@mail.dlut.edu.cn}
	\And
	Zhihui Wang \\
	School of Software \\
	Dalian University of Technology \\
	Dalian, China \\
	\texttt{zhwang@dlut.edu.cn}
	\And
	Peng Xu \\
	School of Software \\
	Dalian University of Technology \\
	Dalian, China \\
	\texttt{2025826903@qq.com}
	\AND
	Xinru Guo \\
	School of Software \\
	Dalian University of Technology \\
	Dalian, China \\
	\texttt{1103394188@mail.dlut.edu.cn}
	\And
	Xintong Li \\
	School of Software \\
	Dalian University of Technology \\
	Dalian, China \\
	\texttt{lixintong2406@mail.dlut.edu.cn}
	\And
	Kun Lu \\
	School of Software \\
	Dalian University of Technology \\
	Dalian, China \\
	\texttt{lukun@dlut.edu.cn}
	\AND
	Ben Fei\thanks{Corresponding author.} \\
	Department of Information Engineering \\
	The Chinese University of Hong Kong \\
	Hong Kong SAR, China \\
	\texttt{benfei@cuhk.edu.hk}
}
\date{}
\renewcommand{\shorttitle}{UniPolymer}

\hypersetup{
pdftitle={UniPolymer: A Unified Framework for Property Prediction, Structure Recommendation, and Evaluation in Polyimide Design},
pdfsubject={Machine Learning for Polyimide Design and Glass Transition Temperature Modeling},
pdfauthor={Junquan Hu, Zhihui Wang, Peng Xu, Xinru Guo, Xintong Li, Kun Lu, Ben Fei},
pdfkeywords={Polyimide Design, Glass Transition Temperature, Property Prediction, Target-Conditioned Generation, Structure Recommendation}
}

\begin{document}

\maketitle

\begin{abstract}
Designing polyimide structures with specific glass transition temperatures (Tg) is highly challenging. Existing methods primarily focus on target-conditioned generation, lacking an assessment of the consistency between the generated structure and the target properties. This leads to low-quality candidates deviating from the design objective entering subsequent processes, increasing invalid experiments and prolonging the development cycle. To address this issue, we propose UniPolymer, a unified framework for property prediction, target-conditioned generation, candidate evaluation, and structure recommendation in polyimide design and a dataset containing 10066 deduplicated polyimide repeating units with Tg tags (PITg-Curated) was constructed. To improve the consistency between generated candidate structures and the target Tg, UniPolymer first establishes a reliable structure-property relationship mapping through self-supervised chemical semantic learning, structural consistency enhancement, and multi-scale information fusion. Subsequently, the model employs a continuous-discrete joint Tg representation to guide the autoregressive generation of SELFIES. The generated candidate structures are further evaluated using a frozen property predictor and polyimide-specific structural constraints, and ranked according to their deviation from the target Tg, thereby preventing structures deviating from the target from entering the subsequent validation stage. Experimental results show that UniPolymer achieved a property prediction accuracy of \(R^2=0.93\) and a candidate structure evaluation pass rate of 73.79\%, which are 2\% and 1.21\% higher than the best baseline, respectively. Meanwhile, the predicted Tg values of the recommended candidates are in high agreement with the results of molecular dynamics simulations, thereby reducing the number of candidates that enter the high-cost experimental stage.
\end{abstract}


\begin{center}
\small
\textbf{Code:} \url{https://github.com/hjunquan1100/UniPolymer}
\quad
\textbf{Datasets:} \url{https://github.com/hjunquan1100/UniPolymer}
\end{center}

\section {Introduction}
Polyimide (PI) exhibits excellent thermal stability, mechanical properties, and chemical resistance~\citep{yi2020high}, and is widely used in aerospace, flexible electronics, and microelectronic packaging. Their glass transition temperatures govern polymer-chain segmental mobility and the applicable operating-temperature range, therefore, precise control of Tg is crucial for application-specific material design~\citep{liu2020high}. 
Traditional development relies on experience-driven structural combinations and iterative synthesis and testing, requiring screening across a vast chemical space, resulting in high costs and long development cycles~\citep{tran2024design}. Therefore, efficient design necessitates prioritizing candidate structures associated with a given Tg. However, the Tg of polyimides is jointly determined by the local chemical environment, segmental conformation, and other factors~\citep{ma2019role}, which are difficult to fully capture using molecular descriptors, structural fingerprints, or a single sequence representation~\citep{huang2025unified}. 
Existing methods mainly use target properties to control the generated distribution but lack verification of the actual properties of generated candidates. Consequently, many low-relevance structures whose properties deviate from the target may enter subsequent stages, resulting in unnecessary experiments and prolonged material development cycles.

To address the aforementioned limitations, we propose UniPolymer, a unified framework integrating property prediction, target condition generation, candidate evaluation, and structure recommendation. To address the difficulty in maintaining consistency between generated structures and target properties, we employ self-supervised chemical semantic pre-training on large-scale unlabeled general polymer data (Supplementary Material S1) during the prediction phase to obtain polymer structure representation. This representation is then transferred to the PITg-Curated (Supplementary Material S2) dataset to establish a reliable polyimide structure-Tg mapping. To enable the model to more accurately learn the polyimide structure distribution under Tg constraints, we employ a continuous-discrete joint Tg conditional pointing to guide the autoregressive generation of SELFIES~\citep{sahu2026polyt5}, reducing the risk of generating invalid molecular sequences. We then use a predictive model to estimate the Tg of candidate structures and rank them based on target bias and physical constraint rules. Finally, molecular dynamics simulations (Supplementary Material S3) are used to further validate the reliability of the recommended structures, thereby reducing the number of low-relevance candidates entering the experimental validation process. The main contributions of this paper are as follows: 
\begin{itemize}
    \setlength{\itemsep}{1pt}
    \item We propose UniPolymer, a unified framework for polyimide Tg prediction, target-conditioned generation, candidate evaluation, and structure recommendation. 
    \item We construct the PITg-Curated dataset containing 10066 deduplicated polyimide repeating units, and achieve discriminative and transferable structural representations from limited labeled material data through polymer semantic pre-training, structural consistency enhancement, and multi-scale information fusion.
    \item We propose a generation strategy for the continuous-discrete joint Tg condition and jointly recommend and rank candidates based on predicted Tg deviations and polyimide-specific structural constraints.
\end{itemize}

\section{Related Work}
\paragraph{Polymer Property Prediction.}
With the rapid development of machine learning and generative artificial intelligence in materials science, data-driven methods have gradually become important tools for understanding polymer structure-property relationships and accelerating candidate material design~\citep{patra2021data}. Early polymer property prediction typically relied on molecular descriptors, structural fingerprints, and quantitative structure-property relationship models, establishing a mapping between polymer structure and target properties through artificially defined structural features~\citep{zhao2023review}.
In polyimide Tg prediction, the combination of molecular dynamics simulations and machine learning has been used to analyze the relationship between structure and thermal properties~\citep{wen2020determination}. Pre-training of graph convolutional networks based on large-scale combinatorial structures and fine-tuning with experimental data has alleviated the problem of limited labeled Tg data~\citep{volgin2022machine}. Building on this, graph neural networks further enhance the modeling ability of structure-Tg relationships and support the screening of target candidates from a large-scale combinatorial space ~\citep{qiu2024heat}.
\paragraph{Polymer Representation Learning.}
To reduce reliance on artificial features, self-supervised learning and chemical language models have begun to be used in polymer representation learning. Transformer-based masked language modeling can extract transferable structural representations from large-scale unlabeled polymer sequences, while modeling polymer structures as a chemical language enables the learning of dense representations applicable to various property prediction tasks ~\citep{kuenneth2023polybert}. These representation learning methods improve the accuracy of property prediction and provide more reliable structure-property relationship knowledge for property-oriented structural design.
\paragraph{Search-Based Polymer Design.}
Based on predictive models, genetic algorithms can iteratively search for polymers that satisfy target properties through crossover and mutation of structural fragments ~\citep{kim2021polymer}, and large-scale virtual screening can also identify target Tg candidates from combinatorially generated polyimide structures ~\citep{qiu2023design}.
However, the exploration scope of search and screening methods is usually limited by predefined structural fragments, combinatorial rules, or candidate databases~\citep{du2024machine}. 

Despite advancements in polymer property prediction, representation learning, and property-guided design, existing methods often lack unified modeling for Tg bias in candidate generation, polyimide-specific structural constraints, and candidate recommendation ranking. UniPolymer addresses this by integrating structure-Tg relationship learning, continuous-discrete condition generation, candidate evaluation, and molecular dynamics validation. It constructs a complete polyimide design pipeline guided by the target Tg, encompassing structure-property relationship learning, target condition generation, candidate screening, and physical validation.

\section{Methodology}
\subsection{Overall Framework}
\begin{figure*}[!htbp]
    \centering
    \includegraphics[width=0.97\textwidth]{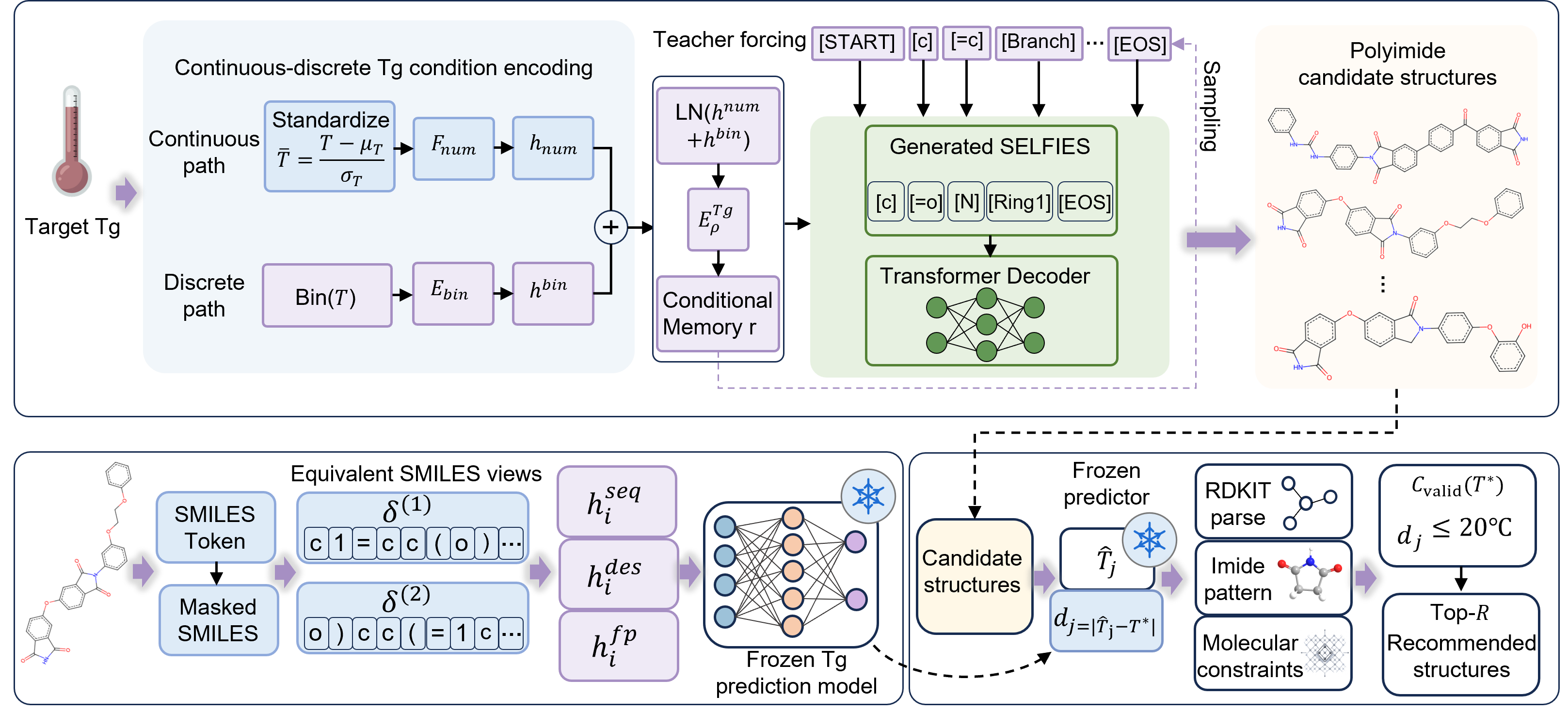}
    \caption{Overview of UniPolymer. By jointly screening based on property consistency and structural rationality, a unified design process is achieved from target Tg input to the recommendation of highly target-relevant structures.}
    \label{fig:framework}
\end{figure*}

UniPolymer first learns the mapping between structure and Tg from polyimide structure-property relationship data. After completing the property model training, it freezes the parameters of the structure encoder and Tg regressor. Then, it explores the candidate structure space of polyimide with the target Tg as a condition~\citep{liu2023high}, and uses the frozen property prediction model to estimate the generated candidate Tg. Combining the predicted Tg bias and polyimide structure constraints, it gives the screening and ranking of high-quality candidates (Figure~\ref{fig:framework}). Finally, it further verifies the reliability of the recommendation results through molecular dynamics simulation calculations~\citep{li2009molecular}. The prediction and generation stages employ different molecular representations and encoding modules. In the prediction stage, each polyimide structure is represented as a SMILES sequence and encoded by the structure encoder \(E_{\phi}\). The resulting sequence representation is fused with RDKit molecular descriptors and Morgan fingerprints, and the fused representation is mapped to a Tg estimate by the regressor \(P_\omega\). In the generation stage, the target Tg is encoded by continuous and discrete condition paths and further transformed by the Tg condition encoder \(E_{\rho}^{\mathrm{Tg}}\) into conditional memory. Conditioned on this memory, the Transformer decoder \(D_{\theta}\) autoregressively generates SELFIES sequences, which are then decoded into candidate molecular structures. Each candidate is subsequently converted into the inputs required by the frozen property predictor and re-encoded for Tg estimation. The relevant formulas will be given in subsequent sections.

Given an unlabeled polymer pretraining dataset \(\mathcal{D}_{\mathrm{pre}}=\{\mathbf{s}_m^{\mathrm{pre}}\}_{m=1}^M\), where \( \mathbf{s}_m^\mathrm{pre}\) denotes the SMILES sequence of the \(m\)-th polymer structure, and \(M\) is the number of pretraining samples. UniPolymer first performs self-supervised chemical-semantic pretraining on \(\mathcal{D}_{\mathrm{pre}}\) to obtain the pre-training parameters of the polymer structure encoder. Then, the pre-trained encoder is transferred to the polyimide structure–property dataset \(A=\{(x_i, T_i)\}_{i=1}^N\), where \(x_i\) and \(T_i\) represent the \(i\)-th polyimide repeating unit and its glass transition temperature, respectively, and \(N\) is the number of samples. UniPolymer learns the Tg-aware structure representation and structure-property relationship mapping from \(A\), \(\hat{T}_i=P_\omega\left(h_i\right)\), where \(\hat{T}_i\) represents the model's predicted Tg value for structure \(x_i\), \(P_\omega\) represents the Tg regressor, and \(\mathbf h_i\) is the multi-scale structure representation, the specific construction process of which will be introduced later. Given a target glass transition temperature \(T^{*}\), the conditional generative model generates a candidate set \(\mathcal{C}(T^*)=\{\widetilde{x}_j\}_{j=1}^J\), where \(\tilde{x}_{j}\) represents the \(j\)-th candidate structure, and \(J\) is the number of candidates. Subsequently, UniPolymer performs Tg estimation and structural verification on each candidate, and prioritizes recommending polyimide structures that are more consistent with the target properties. The property predictor remains frozen throughout candidate generation and evaluation, ensuring that all generated structures are assessed using the same structure--Tg mapping learned from experimental data. This separation prevents the evaluation criterion from changing with the generative model and provides a consistent basis for comparing candidates generated under different target conditions.

\subsection{Tg-Aware Structure–Property Learning}
UniPolymer progressively transfers from general-polymer pretraining~\citep{martin2023emerging} to the polyimide domain through three stages: chemical-semantic pretraining on unlabeled polymers, structure-consistency adaptation on PITg-Curated, and multi-scale supervised Tg regression. This strategy combines transferable polymer knowledge with polyimide-specific structural and property supervision.

In the prediction model stage, we employ a progressive Tg-aware representation learning strategy that transfers from general polymer pre-training~\citep{martin2023emerging} to the polyimide domain. This process includes three stages: general polymer chemical semantic pre-training, polyimide structural consistency enhancement, and multi-scale Tg regression. First, the model learns transferable chemical semantic representations through self-supervised chemical semantic pre-training on large-scale unlabeled polymer structures. Subsequently, structural consistency learning and supervised property regression are performed on the PITg-Curated dataset to construct an efficient model of the polyimide structure-Tg relationship.

\subsubsection{Chemical Semantic Pretraining.}
For the \(m\)-th structure in the general polymer pre-training dataset, it is first converted into a tokenized SMILES sequence \(\mathbf{s}_m^{\mathrm{pre}}=\begin{pmatrix}s_{m, 1}^{\mathrm{pre}}, s_{m, 2}^{\mathrm{pre}},\ldots,s_{m, L_m}^{\mathrm{pre}}\end{pmatrix}\), \(s_{m, l}^{\mathrm{pre}}\) represents the \(l\)-th token in the sequence, and \(L_m\) represents the length of the sequence. Then, UniPolymer performs masked language modeling for self-supervised pre-training~\citep{irwin2022chemformer}. For \(s_{m}^{\mathrm{pre}}\), the mask position set \(\mathcal{M}_m\) is randomly sampled, and the corresponding tokens are replaced with a mask token, resulting in the perturbation sequence \(\mathbf{s}_m^{\mathrm{mask}}\). The encoder learns the contextual chemical semantics in the polymer sequence by recovering the original tokens at the masked positions~\citep{ross2022large}, and its training objective loss is defined as:

\begin{equation}
\label{eq:mlm_loss}
\mathcal{L}_{\mathrm{MLM}}=-\sum_{m=1}^M\sum_{l\in\mathcal{M}_m}\log p_{\phi^{\mathrm{pre}}}\left(s_{m, l}^{\mathrm{pre}}\mid\mathbf{s}_m^{\mathrm{mask}}\right).
\end{equation}
Here, \(p_{\phi^{\mathrm{pre}}}\) is the token distribution predicted by the Transformer encoder \(E_{\phi^{\mathrm{pre}}}\). Recovering the masked tokens enables the encoder to learn functional-group combinations and chain-connectivity patterns, thereby providing transferable chemical priors for Tg modeling with limited labeled data~\citep{li2021mol}.

\subsubsection{Structure-Consistent Representation Enhancement.}
The pretrained encoder is transferred to PITg-Curated for structure-consistency learning. Because one polyimide structure can have multiple equivalent SMILES representations~\citep{arus2019randomized}, differences in sequence notation may cause representation shifts. UniPolymer therefore constructs two randomized SMILES views, \(\mathbf{s}_i^{(1)}\) and \(\mathbf{s}_i^{(2)}\), for each structure and processes them using a shared encoder \(E_\phi\) and projection head \(g_\psi\):
\begin{equation}
\label{eq:normalized_representation}
\mathbf{z}_i^{(a)}
=
\frac{
g_{\psi}\left(E_{\phi}(\mathbf{s}_i^{(a)})\right)
}{
\left\|
g_{\psi}\left(E_{\phi}(\mathbf{s}_i^{(a)})\right)
\right\|_2
}
\quad a\in\{1,2\},
\end{equation}
\(\mathbf{z}_i^{(a)} \) represents the normalized representation corresponding to the \( a \)-th equivalent view. For a training batch containing \( B \) samples, a symmetric contrast objective is used:
\begin{equation}
\label{eq:contrastive_loss}
\begin{aligned}
\mathcal{L}_{\mathrm{con}}
=
-\frac{1}{2B}\sum_{i=1}^{B}
\Bigg[
&\log
\frac{
\exp\left(
\operatorname{sim}
(\mathbf{z}_i^{(1)}, \mathbf{z}_i^{(2)})/\tau
\right)
}{
\sum_{j=1}^{B}
\exp\left(
\operatorname{sim}
(\mathbf{z}_i^{(1)}, \mathbf{z}_j^{(2)})/\tau
\right)
}
+
\log
\frac{
\exp\left(
\operatorname{sim}
(\mathbf{z}_i^{(2)}, \mathbf{z}_i^{(1)})/\tau
\right)
}{
\sum_{j=1}^{B}
\exp\left(
\operatorname{sim}
(\mathbf{z}_i^{(2)}, \mathbf{z}_j^{(1)})/\tau
\right)
}
\Bigg].
\end{aligned}
\end{equation}
Here, \(\operatorname{sim}(\cdot,\cdot)\) denotes cosine similarity and \(\tau\) is the temperature coefficient. The objective pulls equivalent SMILES representations together while separating different structures in the batch, transferring pretrained chemical semantics to the polyimide domain and reducing sensitivity to SMILES notation~\citep{lee2025simson}.
\subsubsection{Multi-Scale Tg Regression.}
To compensate for the insufficient coverage of local and global structural information by single sequence representations, UniPolymer combines the Transformer sequence representation \(\mathbf{h}_i^{\mathrm{seq}}=E_\phi(\mathbf{s}_i)_{[\mathrm{CLS}]}\), RDKit molecular descriptors \(\mathbf{h}_i^{\mathrm{des}}=F_{\mathrm{des}}(x_i)\), and Morgan fingerprints \(\mathbf{h}_i^{\mathrm{fp}}=F_{\mathrm{fp}}(x_i)\)~\citep{rogers2010extended}. These features encode sequence context, global molecular properties, and local substructure patterns, respectively.

The three types of structural information are then fused by vector concatenation along the feature dimension:
\begin{equation}
\label{eq:multiscale_fusion}
\mathbf{h}_i
=
\begin{bmatrix}
\mathbf{h}_i^{\mathrm{seq}}; 
\mathbf{h}_i^{\mathrm{des}}; 
\mathbf{h}_i^{\mathrm{fp}}
\end{bmatrix}.
\end{equation}
Subsequently, the regressor \(P_{\omega}\) gives the predicted Tg of the structure \(x_i\), and the property regression stage uses a sample-wise weighted Huber loss:
\begin{equation}
\label{eq:tg_loss}
\mathcal{L}_{\mathrm{Tg}}
=
\frac{1}{B}
\sum_{i=1}^{B}
w_i\,
\ell_{\mathrm{Hub}}
\left(
\widehat{T}_i, T_i
\right).
\end{equation}
Here, \(w_i\) is the weight of the \(i\)-th sample, and \(\ell_{\mathrm{Hub}}()\) is the Huber error function. Sequence semantics, overall molecular properties, and local substructure information work together to form a Tg-aware representation and establish a structure-Tg mapping~\citep{shen2024complementary}. After training, the encoder \(E_{\phi}\), feature fusion module, and regressor \(P_{\omega}\) are all frozen and used only to generate candidate Tg estimates.
\subsection{Target-Conditioned Polyimide Generation}
After establishing the structure-Tg mapping, UniPolymer explores the polyimide structure space conditioned on the target Tg~\citep{jiang2024property}. To balance continuous numerical relationships with temperature-level differences, we construct a continuous-discrete cooperative Tg conditional representation and use it for SELFIES autoregressive generation. For the Tg value \(T_i\), it is first standardized using the mean \(\mu_T\) and standard deviation \(\sigma_T\) of the training set: \(\overline{T}_i=\frac{T_i-\mu_T}{\sigma_T}\). Then, the continuous path obtains the continuous condition vector \(\mathbf{h}_i^\mathrm{num}=F_\mathrm{num}(\overline{T}_i)\) via the non-linear mapping \(F_{\mathrm{num}}\), where \(\mathbf{h}_i^{\mathrm{num}}\) represents the continuous Tg condition vector. Meanwhile, the Tg range is divided into \(K_T\) temperature intervals, and the corresponding discrete temperature interval index is obtained via \(\operatorname{Bin}()\):
\begin{equation}
\label{eq:temperature_bin}
b_i
=
\operatorname{Bin}(T_i),
\quad
b_i\in\{0, 1, \ldots, K_T-1\}.
\end{equation}
The discrete path embeds the temperature-bin index as \(\mathbf{h}_i^\mathrm{bin}\) using \(E_{\mathrm{bin}}\). The continuous and discrete condition vectors are then fused in a shared latent space~\citep{pinheiro2022smiclr}:
\begin{equation}
\label{eq:tg_condition_fusion}
\mathbf{h}_i^{\mathrm{Tg}}
=
\operatorname{LN}
\left(
\mathbf{h}_i^{\mathrm{num}}
+
\mathbf{h}_i^{\mathrm{bin}}
\right),
\end{equation}
\(\mathrm{LN()}\) represents the layer normalization operation, and \(\mathbf{h}_i^{\mathrm{Tg}}\) represents the fused Tg conditional representation. Finally, the conditional encoder \(E_{\rho}^{\mathrm{Tg}}\) further encodes it into conditional memory:
\begin{equation}
\label{eq:conditional_memory}
\mathbf{r}_i
=
E_{\rho}^{\mathrm{Tg}}
\left(
\mathbf{h}_i^{\mathrm{Tg}}
\right).
\end{equation}
Here, continuous paths preserve the fine-grained numerical relationship of Tg~\citep{born2023regression}, while discrete paths characterize its temperature range information, together, they provide complementary target conditions.
\subsubsection{Conditional SELFIES Autoregressive Generation.}
To improve the chemical robustness of the generated sequences, the polyimide structure is first converted into SELFIES~\citep{krenn2020self, nigam2021beyond}, the polyimide structure is first converted into SELFIES. The \(i\)-th structure is represented as \(\mathbf{u}_i=(u_{i, 1}, u_{i, 2}, \ldots, u_{i, S_i})\), where \(S_i\) is the sequence length. Given the Tg conditional memory \(\mathbf{r}_i\), the conditional Transformer Decoder \(D_{\theta}\) performs autoregressive modeling on the structure sequence:
\begin{equation}
\label{eq:autoregressive_generation}
p_{\theta}
\left(
\mathbf{u}_i\mid T_i
\right)
=
\prod_{t=1}^{S_i}
p_{\theta}
\left(
u_{i, t}\mid u_{i<t}, \mathbf{r}_i
\right),
\end{equation}
where \(p_{\theta}\) is the conditional probability distribution with parameter \(\theta\), and \(u_{i<t}\) represents the SELFIES tokens before position \(t\). Teacher forcing is used during the training phase~\citep{williams1989learning}. Let \(\mathcal{V}\) denote the SELFIES vocabulary and \(\Omega\) the set of all non-padding positions in a batch. The generation loss is then defined as:
\begin{equation}
\label{eq:generation_loss}
\mathcal{L}_{\mathrm{gen}}
=
-\frac{1}{|\Omega|}
\sum_{(i, t)\in\Omega}
\sum_{v\in\mathcal{V}}
q_{i, t, v}
\log p_{\theta}
\left(
v\mid u_{i<t},\mathbf{r}_i
\right),
\end{equation}
where \(v\) represents any token in the vocabulary, and \(q_{i, t, v}\) represents the label smoothing supervision probability:
\begin{equation}
\label{eq:label_smoothing}
q_{i, t, v}
=
\begin{cases}
1-\varepsilon,
& v=u_{i, t} \\[2pt]
\dfrac{\varepsilon}{|\mathcal{V}|-1}
& v\neq u_{i, t}
\end{cases},
\end{equation}
\(\varepsilon\) represents the label smoothing coefficient. During the inference phase, the target Tg undergoes continuous-discrete conditional encoding to obtain \(\mathbf{r}_i\). The model progressively generates SELFIES from the initial token and uses nucleus sampling to obtain candidates that possess both target relevance and structural diversity~\citep{holtzman2019curious}, ultimately decoding them into polyimide structures.

\subsection{Recommendation and Evaluation}
To ensure consistency between the generated candidates and the target Tg, UniPolymer uses a frozen prediction model to evaluate the properties of the candidates \(\tilde{x}_{j}\) and defines the target bias as \(d_j=\left|\widehat{T}_j-T^*\right|\), where a smaller \(d_j\) indicates a greater consistency between the candidate and the target Tg. Simultaneously, structural verification is performed on the candidates:

\begin{equation}
\label{eq:structural_verification}
\chi(\widetilde{x}_j)
=
\chi_{\mathrm{parse}}(\widetilde{x}_j)
\chi_{\mathrm{imide}}(\widetilde{x}_j)
\chi_{\mathrm{mol}}(\widetilde{x}_j).
\end{equation}

Here, \(\chi(\widetilde{x}_j)\in\{0,1\}\) indicates whether the candidate has passed structural verification, \(\chi_{\mathrm{parse}}\) indicates the RDKit parsability check, \(\chi_{\mathrm{imide}}\) represents the imide skeleton check, \(\chi_{\mathrm{mol}}\) represents the molecular property check, the judgment process is as follows:

\begin{equation}
\label{eq:molecular_constraints}
\begin{aligned}
\chi_{\mathrm{mol}}(x)
={}&
\mathbb{I}
\left(
150\leq M_{\mathrm{W}}(x)\leq 2000
\right)
\\
&\times
\mathbb{I}
\left(
10\leq n_{\mathrm{H}}(x)\leq 150
\right)
\mathbb{I}
\left(
U(x)=0
\right)
\end{aligned}.
\end{equation}

Among them, \(M_{\mathrm{W}}(x)\), \(n_{\mathrm{H}}(x)\), and \(U(x)\) represent the detection results of molecular weight, number of heavy atoms, and unstable groups, respectively. The target-related effective candidate set is defined as:

\begin{equation}
\label{eq:valid_candidate_set}
\mathcal{C}_{\mathrm{valid}}(T^*)
=
\left\{
\widetilde{x}_j\in\mathcal{C}
\;\middle|\;
d_j\leq\delta_T, \,
\chi(\widetilde{x}_j)=1
\right\}.
\end{equation}

Here, \(\delta_T=20\,^\circ\mathrm{C}\) is the maximum allowable absolute deviation from the target Tg. UniPolymer sorts the candidates in \(\mathcal{C}_{\mathrm{valid}}(T^*)\) in ascending order of \(d_j\) and returns the top \(R\) recommended structures:
\begin{equation}
\label{eq:recommended_candidate_set}
\mathcal{R}_{R}(T^*)
=
\operatorname*{arg\,top\mathcal{R}}_{
\widetilde{x}_j
\in
\mathcal{C}_{\mathrm{valid}}(T^*)
}
\left(
-d_j
\right).
\end{equation}
\(\mathcal{R}_{R}(T^{*})\) represents the recommended candidate set corresponding to the target \(T^{*}\). Through this procedure, UniPolymer performs candidate-property evaluation, structural screening, and target-relevant structure recommendation.
\section{Experiments and Results}
\subsection{Datasets and Settings}
This paper constructs a polyimide structure-glass transition temperature dataset, PITg-Curated, containing 10066 deduplicated polyimide repeating units and their Tg labels. The cleaned Tg ranges from 19 to 730 °C. The dataset is divided into training, validation, and test sets at a ratio of 8:1:1 and represented using SMILES and SELFIES, respectively, for Tg-aware representation learning and target condition generation.
In the property prediction task, MAE, RMSE, and \(R^2\) are used for evaluation~\citep{chicco2021coefficient}. The generation task is evaluated from two perspectives: structural validity and target-property consistency. Structural validity includes RDKit parsability and compliance with polyimide-specific rules covering the imide skeleton, molecular weight, heavy-atom count, and potentially unstable groups. Target-property consistency is evaluated using the frozen Tg predictor. The results were evaluated using Target MAE and Hit@20. The proportion of candidates that simultaneously satisfy both structural constraints and \(|\widehat T-T^*|\leq20^\circ\mathrm{C}\) was defined as the final recommendation pass rate (Valid  Hit@20). Furthermore, the top 100 candidates under each target Tg condition were selected for molecular dynamics simulations, and the results were compared with model predictions to verify the reliability of the property assessment and recommendation results.

We implement UniPolymer in PyTorch 2.5.1 with Python 3.12 and CUDA 12.4, and conduct all experiments on an NVIDIA A40 GPU, and the source code is publicly available. The model uses the AdamW optimizer with an initial learning rate of \(1\times10^{-4}\), a weight decay of \(1\times10^{-2}\), and a label smoothing coefficient of 0.1.
\subsection{Results and Analysis}
\subsubsection{Tg Prediction Performance and Error Analysis.}
\begin{figure}[!htbp]
    \centering
    \includegraphics[width=\columnwidth]{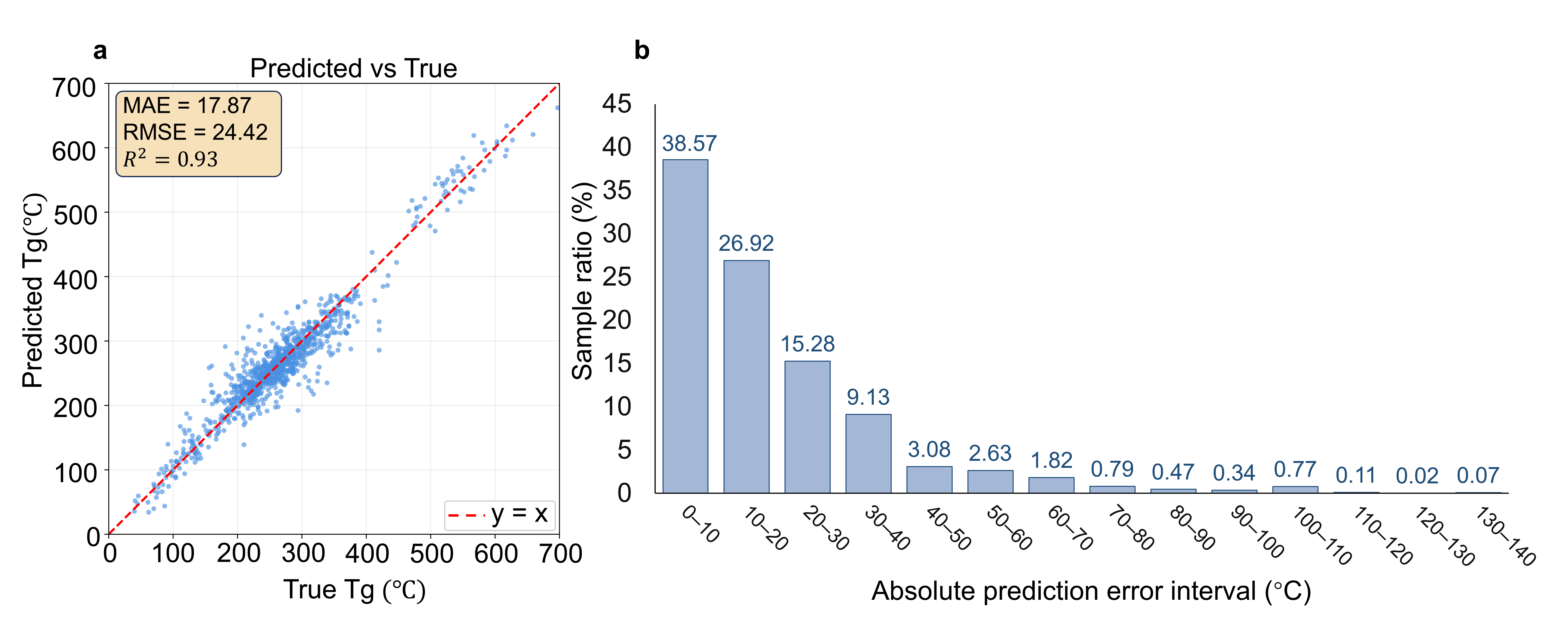}
    \caption{UniPolymer's prediction performance of polyimide Tg on the test set. The predicted values and experimental values generally show a high consistency.}
    \label{fig:prediction}
\end{figure}

\begin{table}[!htbp]
    \centering
    \setlength{\tabcolsep}{2.2pt}
    \renewcommand{\arraystretch}{1.05}
    \begin{tabularx}{\columnwidth}{@{}>{\raggedright\arraybackslash}Xccc@{}}
        \toprule
        Model
        & $R^2$ $\uparrow$
        & \makecell{MAE ($^{\circ}\mathrm{C}$) $\downarrow$}
        & \makecell{RMSE ($^{\circ}\mathrm{C}$) $\downarrow$} \\
        \midrule
        DNN
        & 0.85
        & 20.09
        & 27.60 \\

        ANN
        & 0.78
        & 23.88
        & 33.09 \\

        CatBoost
        & 0.90
        & 18.58
        & \textbf{23.06} \\

        Importance-Transformer
        & 0.84
        & 20.61
        & 27.64 \\

        TransPolymer
        & 0.89
        & 18.77
        & 27.86 \\

        polyBARTguoxinru1120@gmail.com
        & 0.91
        & 18.14
        & 38.41 \\

        LLaMA-3-8B
        & 0.88
        & 20.53
        & 58.13 \\

        \textbf{UniPolymer}
        & \textbf{0.93}
        & \textbf{17.87}
        & 24.42 \\
        \bottomrule
    \end{tabularx}
    \caption{Tg prediction performance on the PITg-Curated test set.}
    \label{tab:model_comparison}
\end{table}
On the test set (Figure~\ref{fig:prediction}(a)), UniPolymer achieves \(R^2=0.93\),  \(\mathrm{MAE}=17.87\,^\circ\mathrm{C}\), and  \(\mathrm{RMSE}=24.42\,^\circ\mathrm{C}\), providing a reliable structure--property mapping for candidate evaluation.
Figure~\ref{fig:prediction}(b) further shows that 65.49\% of the test samples fall within 20°C of the experimental Tg. As summarized in Table~\ref{tab:model_comparison}, UniPolymer outperforms the representative baselines in terms of \(R^2\) and MAE.

\FloatBarrier
\subsubsection{Chemical Space Generation and Structural Quality Analysis.}
This paper uses Morgan fingerprinting to perform UMAP dimensionality reduction analysis on the distribution relationship between generated candidate polyimides and real polyimides in chemical space.
\begin{figure}[!htbp]
    \centering
    \includegraphics[width=\columnwidth]{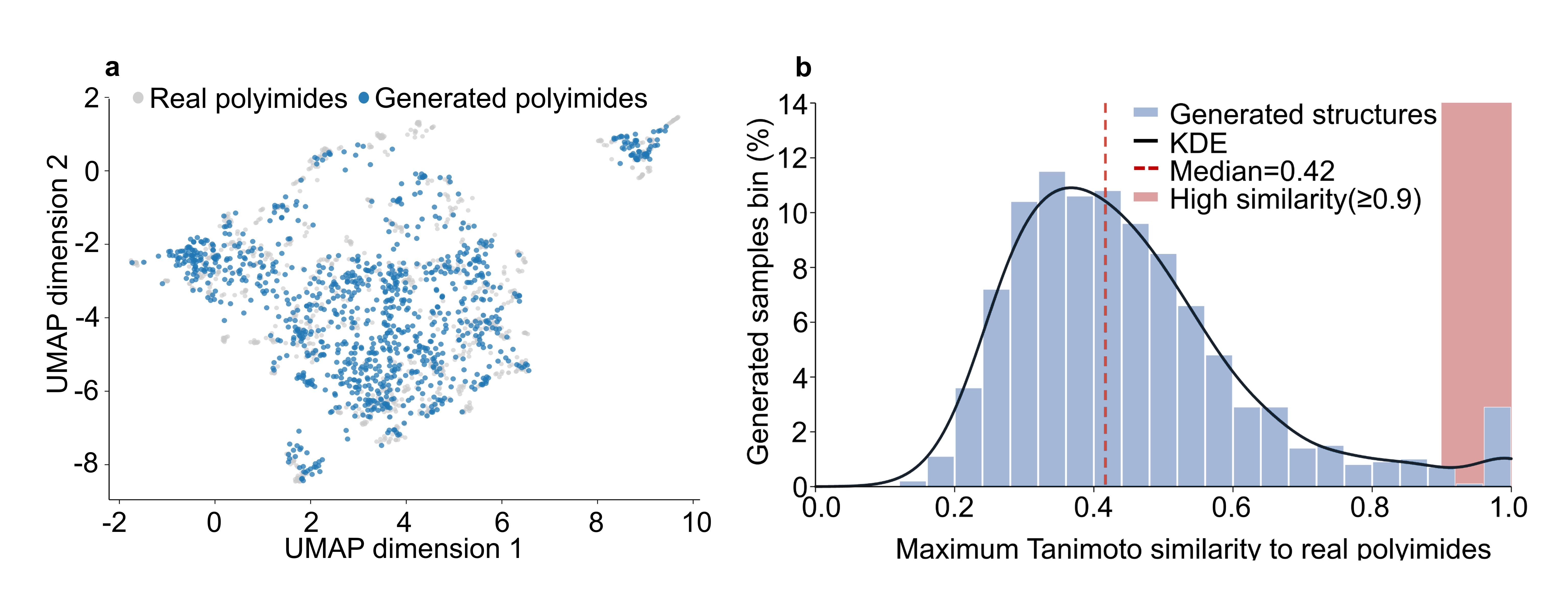}
    \caption{Generated and real polyimide structures are compared in terms of chemical-space distributions and structural similarities. UniPolymer captures the principal structural characteristics of real polyimides while expanding the explored chemical space.}
    \label{fig:chemical_space}
\end{figure}

Figure~\ref{fig:chemical_space}(a) shows that the generated structures largely overlap with real polyimides in the principal chemical-space regions, indicating that the model learns the dominant distribution of the training data rather than producing structurally isolated samples. Figure~\ref{fig:chemical_space}(b) provides a complementary structure-level analysis by reporting the maximum Tanimoto similarity~\citep{bajusz2015tanimoto} between each generated structure and its nearest neighbor in the real dataset. The median similarity of 0.42 suggests that the candidates retain recognizable polyimide characteristics without simply reproducing known structures. Taken together, the two analyses show that UniPolymer preserves the underlying chemical distribution while extending it toward less explored regions. Nevertheless, chemical-space coverage does not necessarily imply chemical validity or compliance with polyimide-specific requirements. We therefore evaluate all generation baselines on 10000 candidates using RDKit validity, imide-skeleton constraints, molecular-property ranges, structural stability, novelty, plus the overall pass rate. As shown in Table~\ref{tab:structure_validity}, UniPolymer reaches 100\% RDKit validity, a 77.63\% imide-skeleton pass rate, 65.54\% novelty, with an overall pass rate of 73.79\%. This final metric exceeds the strongest baseline by 1.21 percentage points, showing that the expansion of chemical space is achieved without sacrificing the structural quality of generated candidates.
\begin{table*}[!htbp]
    \centering
    \footnotesize
    \setlength{\tabcolsep}{1.2pt}
    \renewcommand{\arraystretch}{1.08}

    \begin{tabular*}{\textwidth}{
        @{\extracolsep{\fill}}
        l
        c
        c
        c
        c
        c
        c
        c
        @{}
    }
        \toprule

        \multirow{2}{*}{Model}
        & \multicolumn{1}{c}{Chemical Validity}
        & \multicolumn{1}{c}{Structural Validity}
        & \multicolumn{3}{c}{Molecular Property Validity}
        & \multirow{2}{*}{Novelty}
        & \multirow{2}{*}{Valid} \\

        \cmidrule(lr){2-2}
        \cmidrule(lr){3-3}
        \cmidrule(lr){4-6}

        & RDKit
        & Imide skeleton
        & Molecular weight range
        & Heavy atom count
        & Stable groups
        & & \\

        \midrule

        CharRNN
        & 69.57
        & 58.43
        & 93.27
        & 91.60
        & 84.72
        & 38.96
        & 29.38 \\

        GraphINVENT
        & 79.86
        & 64.37
        & 95.14
        & 93.82
        & 87.63
        & 47.54
        & 40.16 \\

        REINVENT-RL
        & 94.62
        & 73.84
        & 97.67
        & 89.93
        & 92.46
        & 61.27
        & 61.02 \\

        PolyTAO-style
        & \textbf{100.00}
        & 75.96
        & \textbf{99.93}
        & 98.17
        & \textbf{94.92}
        & 63.86
        & 72.58 \\

        PoGE
        & 96.75
        & 68.74
        & 96.28
        & 95.47
        & 83.69
        & 57.38
        & 53.97 \\

        \textbf{UniPolymer}
        & \textbf{100.00}
        & \textbf{77.63}
        & 99.91
        & \textbf{100.00}
        & 94.38
        & \textbf{65.54}
        & \textbf{73.79} \\

        \bottomrule
    \end{tabular*}

    \caption{Structural and molecular-property validity of generated candidates (\%).}
    \label{tab:structure_validity}
\end{table*}

\FloatBarrier
\subsubsection{Evaluation of Target-Tg Conditional Response and Property Consistency.}
In addition to structural quality, this paper further evaluates the responsiveness of the generated candidate structures to the target Tg. Under four target Tg conditions \((T^*=100^\circ\mathrm{C}, 200^\circ\mathrm{C}, 300^\circ\mathrm{C}, 400^\circ\mathrm{C})\), we generate 1000 candidates and report Target MAE, Hit@20, and the final recommendation pass rate. Model predictions are further compared with molecular-dynamics results for the top 100 candidates per target.

\begin{figure*}[!htbp]
    \centering
    \includegraphics[width=\textwidth]{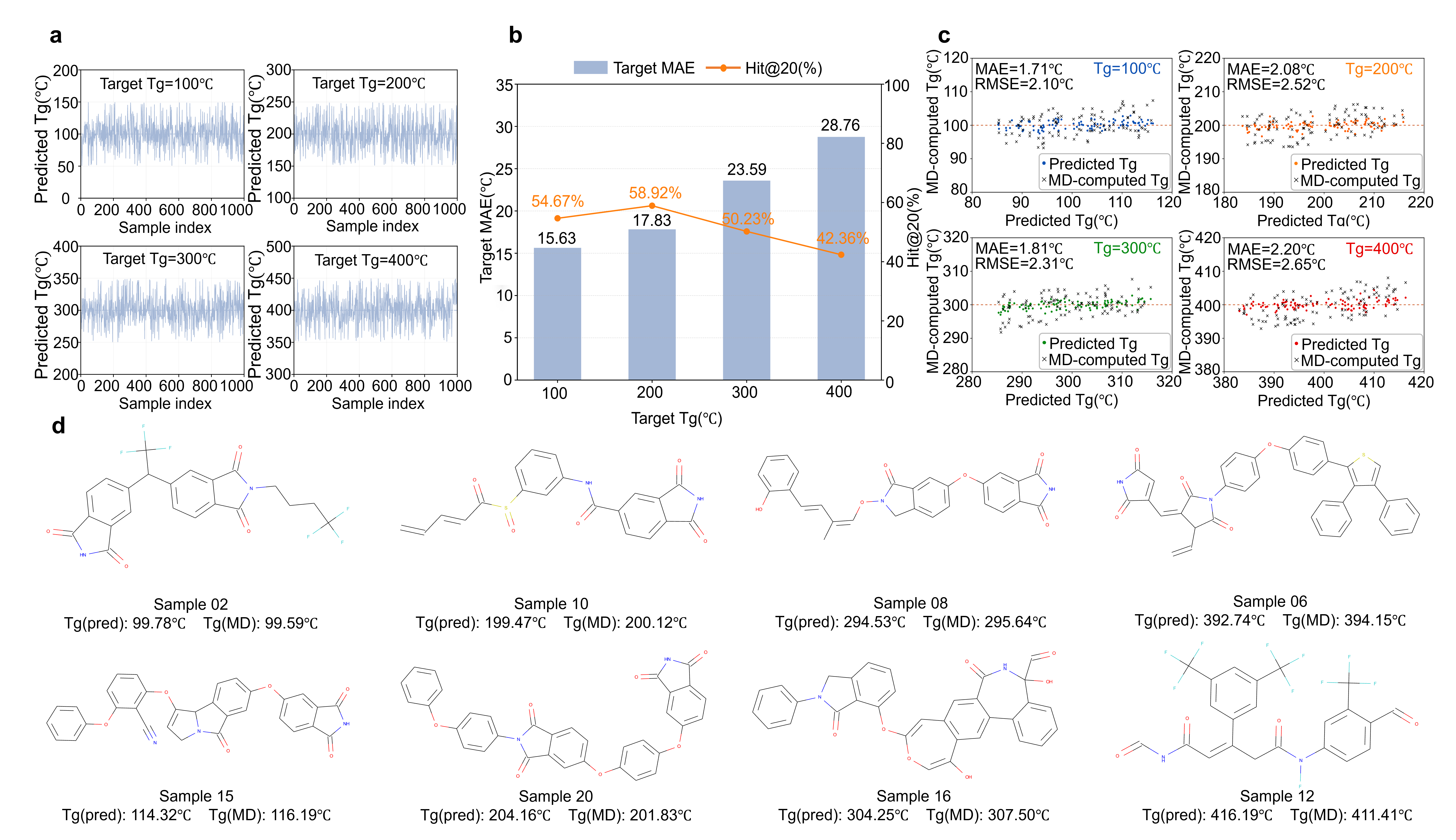}
    \caption{Target-Tg-driven polyimide structure exploration:
(a) predicted Tg distributions; (b) Target MAE and Hit@20;
(c) agreement between model predictions and MD estimates;
(d) representative recommended structures.}
    \label{fig:target_generation}
\end{figure*}

As shown in Figure~\ref{fig:target_generation}(a), the predicted Tg distributions shift toward the corresponding temperature ranges as the target Tg increases and remain concentrated around the target values. The Target MAE values under the four target conditions are 15.63, 17.83, 23.59 and 28.76~$^\circ$C, with Hit@20 values of 54.67\%, 58.92\%, 50.23\% and 42.36\%, respectively (Figure~\ref{fig:target_generation}(b)). To verify the reliability of the recommendation results, molecular dynamics simulations were performed for the top 100 candidates recommended in each target-temperature range (Figure~\ref{fig:target_generation}(c)). The comparison between the model predictions and MD estimates yields average MAE and RMSE values of 1.95~$^\circ$C and 2.40~$^\circ$C, respectively. Most candidates have simulated Tg values close to the corresponding target temperatures, while Figure~\ref{fig:target_generation}(d) presents representative candidate structures from each temperature range.

\FloatBarrier
\subsection{Ablation Experiment}
We ablate the Tg-aware representation and target-condition generation modules separately. 

\begin{table}[!htbp]
    \centering
    \setlength{\tabcolsep}{2.1pt}
    \renewcommand{\arraystretch}{1.04}
    \begin{tabularx}{\columnwidth}{@{}>{\raggedright\arraybackslash}Xccc@{}}
        \toprule
        Method
        & $R^2$ $\uparrow$
        & \makecell{MAE ($^{\circ}\mathrm{C}$) $\downarrow$}
        & \makecell{RMSE ($^{\circ}\mathrm{C}$) $\downarrow$} \\
        \midrule
        w/o Chemical Semantic Pretraining
        & 0.86
        & 21.57
        & 26.18 \\

        w/o Structure Consistency
        & 0.89
        & 19.96
        & 25.72 \\

        w/o Molecular Descriptors
        & 0.91
        & 18.12
        & 25.21 \\

        w/o Morgan Fingerprints
        & 0.84
        & 23.42
        & 27.14 \\

        \textbf{UniPolymer (pred)}
        & \textbf{0.93}
        & \textbf{17.87}
        & \textbf{24.42} \\
        \bottomrule
    \end{tabularx}
    \caption{Ablation study of the Tg prediction module.}
    \label{tab:prediction_ablation}
\end{table}
\begin{table}[!htbp]
    \centering
    \setlength{\tabcolsep}{2.2pt}
    \renewcommand{\arraystretch}{1.05}
    \begin{tabularx}{\columnwidth}{@{}>{\raggedright\arraybackslash}Xccc@{}}
        \toprule
        Method
        & Target MAE ($^{\circ}\mathrm{C}$) $\downarrow$
        & Hit@20 (\%) $\uparrow$
        & Valid Hit@20 (\%) $\uparrow$ \\
        \midrule

        Continuous Condition
        & 24.87
        & 49.76
        & 46.37 \\

        Discrete Condition
        & 26.16
        & 48.61
        & 42.92 \\

        \textbf{UniPolymer (gen)}
        & \textbf{21.46}
        & \textbf{51.55}
        & \textbf{50.84} \\

        \bottomrule
    \end{tabularx}
    \caption{Ablation of the target-Tg condition representation.}
    \label{tab:condition_ablation}
\end{table}

Table~\ref{tab:prediction_ablation} shows that each component contributes to prediction performance. Removing Morgan fingerprints causes the largest degradation: \(R^2\) falls to 0.84, whereas MAE rises to 23.42~$^\circ$C, underscoring the importance of local substructure patterns for distinguishing polyimides with different Tg values. Excluding semantic pretraining or structure consistency raises MAE to 21.57 and 19.96~$^\circ$C, respectively. Together, these ablations show that self-supervised chemical knowledge, invariance across equivalent SMILES representations, plus multi-scale structural cues support reliable Tg modeling.

Turning to conditional generation, Table~\ref{tab:condition_ablation} compares the complete condition representation against continuous-only and discrete-only variants, averaged over four target temperatures. Relative to these variants, UniPolymer lowers Target MAE by 3.41 and 4.70~$^\circ$C; Hit@20 increases by 1.79 and 2.94 percentage points; Valid Hit@20 rises by 4.47 and 7.92 percentage points, respectively. These gains suggest that continuous conditions preserve fine-grained numerical variation, whereas discrete conditions encode complementary temperature ranges. Their combination yields more accurate target-conditioned generation.

\FloatBarrier
\section{Conclusion}
This paper presents UniPolymer, a unified framework designed to maintain consistency between generated polyimide candidates and the target Tg through property prediction, conditional generation, candidate evaluation, and structure recommendation. It establishes a reliable structure-Tg mapping through self-supervised chemical semantic learning, structural consistency enhancement, and multi-scale information fusion, and guides SELFIES generation using continuous--discrete Tg conditions.
Candidates are further ranked according to their absolute deviations from the target Tg and their compliance with structural constraints. UniPolymer achieves \(R^2=0.93\) and an MAE of \(17.87~^\circ\mathrm{C}\) for Tg prediction, together with a 73.79\% structural pass rate. The close agreement between model predictions and MD-estimated Tg values further supports the reliability of candidate evaluation. Future research will focus on improving the prediction and generation capabilities for high Tg regions and introducing synthetic feasibility, uncertainty assessment, and wet experimental validation of recommended structures to enhance the reliability and practical application value of recommended candidates.

\section{Acknowledgments}
This work was supported by the National Science and Technology Major Project on New Generation Artificial Intelligence: Research on Intelligent Methods for Scientific Research Oriented to Materials Design (No. 2025ZD0121802), the Data Intelligence Innovation Base Project (No. FPF10120260003), and the Scientific Embodied Multi-Agent System Project (No. FPF10120250008).

\FloatBarrier
\appendix
\section{Appendix}

% Appendix figures are numbered S1, S2, ...
\setcounter{figure}{0}
\renewcommand{\thefigure}{S\arabic{figure}}

% Appendix equations are numbered S1, S2, ...
\setcounter{equation}{0}
\renewcommand{\theequation}{S\arabic{equation}}

\subsection{S1: General-Polymer Pretraining Corpus}

\subsubsection{Data Source and Corpus Construction}

To obtain transferable chemical representations before adapting the
structure encoder to the polyimide domain, we employed the unlabeled
polymer corpus released with TransPolymer~\citep{xu2023transpolymer}, which was constructed from
the PI1M database. PI1M contains approximately one million hypothetical
polymer repeating-unit structures generated from polymer structures
collected from the PoLyInfo database.

Each polymer repeating unit is represented using a polymer SMILES
(P-SMILES) sequence, in which two wildcard atoms (\texttt{*}) indicate
the polymerization sites connecting adjacent repeating units.
Following the data-processing procedure of TransPolymer,
canonicalization was removed, and non-canonical SMILES corresponding
to the same polymer structure were generated for sequence augmentation.
Each original polymer entry was augmented into five SMILES
representations, producing approximately five million unlabeled
polymer sequences.

% ============================================================
% S2
% ============================================================

\subsection{S2: Construction and Representation of the PITg-Curated Dataset}

\subsubsection{Data Collection and Standardization}
\label{appendix:data-collection}
The PITg-Curated dataset contains 10066 deduplicated polyimide
repeating units and their glass transition temperatures. Of these,
2129 records (21.15\%) originated from molecular dynamics simulations
conducted in this paper, while 7,937 records (78.85\%) were collected
from publicly available polymer databases and peer-reviewed literature.

The publicly available data were primarily collected from PoLyInfo~\citep{ishii2024polyinfo},
the Polymer Property Predictor and Database, and related literature.
The repeating-unit structures, Tg values, temperature units, and
measurement methods were manually verified. Records with unclear
structural or property information were removed.

All structures were parsed and normalized using RDKit~\citep{rdkit2026} and converted
into canonical SMILES. Tg values were uniformly expressed in degrees
Celsius, and records with the same canonical SMILES were considered
to represent the same polyimide structure. For structures with multiple
valid Tg labels, the median was used as the final label to reduce the
influence of differences among data sources, testing conditions, and
outliers.

\subsubsection{Statistical Characterization of PITg-Curated}

\begin{figure}[!t]
    \centering
    \includegraphics[width=\columnwidth]{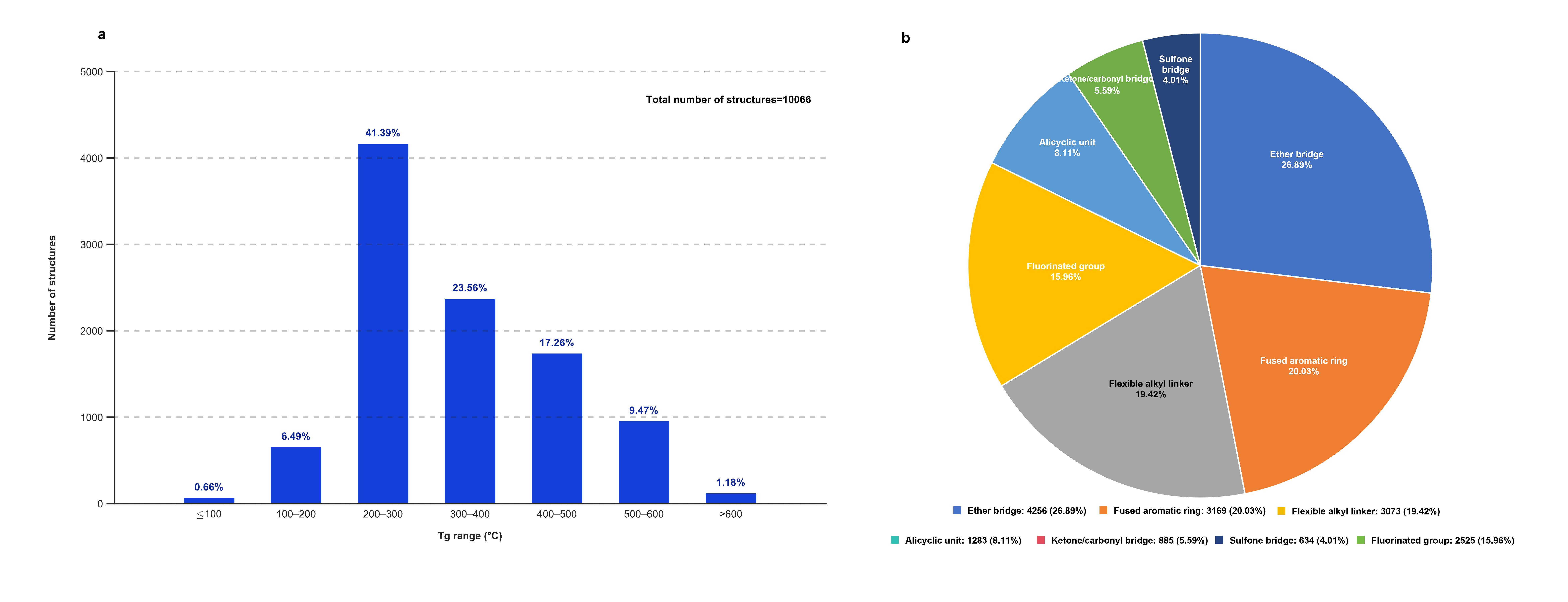}
    \caption{
    Statistical characterization of the PITg-Curated dataset.
    (a) Distribution of polyimide structures across different Tg ranges.
    (b) Composition of representative structural groups in the dataset.
    }
    \label{fig:dataset-statistics}
\end{figure}

The Tg values in PITg-Curated range from
\(19\,^{\circ}\mathrm{C}\) to \(730\,^{\circ}\mathrm{C}\),
with a mean, median, and standard deviation of
\(334.74\,^{\circ}\mathrm{C}\),
\(305.00\,^{\circ}\mathrm{C}\), and
\(112.98\,^{\circ}\mathrm{C}\), respectively.

As shown in Figure~\ref{fig:dataset-statistics}(a), the data are mainly
distributed within the \(200\)--\(500\,^{\circ}\mathrm{C}\) range,
accounting for 82.21\% of all samples, with the
\(200\)--\(300\,^{\circ}\mathrm{C}\) range containing the largest
number of samples. At the same time, the dataset also covers low- and
high-Tg regions, forming a relatively broad temperature distribution
and providing a data foundation for learning polyimide
structure--property relationships across different temperature ranges.

To further characterize the structural composition of the dataset,
we statistically analyzed representative structural features,
including ether bridges, fused aromatic rings, flexible alkyl linkers,
fluorinated groups, alicyclic units, ketone/carbonyl bridges, and
sulfone bridges, based on RDKit substructure matching.

As shown in Figure~\ref{fig:dataset-statistics}(b), PITg-Curated
contains both rigid structures represented by fused aromatic rings
and flexible structures containing ether bridges, flexible alkyl
linkers, and alicyclic units. It also contains various fluorinated and
polar functional groups. These structural features indicate that the
dataset exhibits structural diversity in terms of chain-segment
rigidity, bridging mechanisms, and local chemical environments,
providing a sufficient chemical basis for the model to learn the
complex relationship between polyimide structure and Tg.

\FloatBarrier

% ============================================================
% S3
% ============================================================

\subsection{S3: Molecular Dynamics Simulation and Glass Transition Temperature Calculation}

To supplement the experimental data available in publicly accessible
databases and literature, we further employed all-atom molecular
dynamics simulations to calculate the glass transition temperatures
of selected polyimide structures~\citep{huo2025polyimide,suter2025rapid}.

The complete calculation procedure included polymer-chain construction,
amorphous periodic model generation, geometry optimization, cyclic
annealing, multi-temperature kinetic equilibration, and
density--temperature fitting. All simulations were performed using Materials Studio. The amorphous
models were constructed using the Amorphous Cell module, while geometry
optimization and molecular dynamics calculations were performed using
the Forcite module. Interatomic interactions were described using the
COMPASS force field, following established atomistic simulation
procedures for polyimide Tg evaluation
~\citep{huo2025polyimide}.
The key steps of the simulation procedure are described below.

\subsubsection{Construction of Polymer Chains and Amorphous Periodic Models}
\label{appendix:amorphous-model}

For each polyimide structure, a corresponding three-dimensional
molecular model was first established based on its repeating unit.
By specifying the first and last atoms of the repeating unit, the
polymerization connection sites between adjacent repeating units were
determined. The repeating units were then sequentially connected along
the polymerization direction to construct a linear homopolymer chain
with a preset chain length.

\begin{figure}[!t]
    \centering
    \includegraphics[width=\columnwidth]{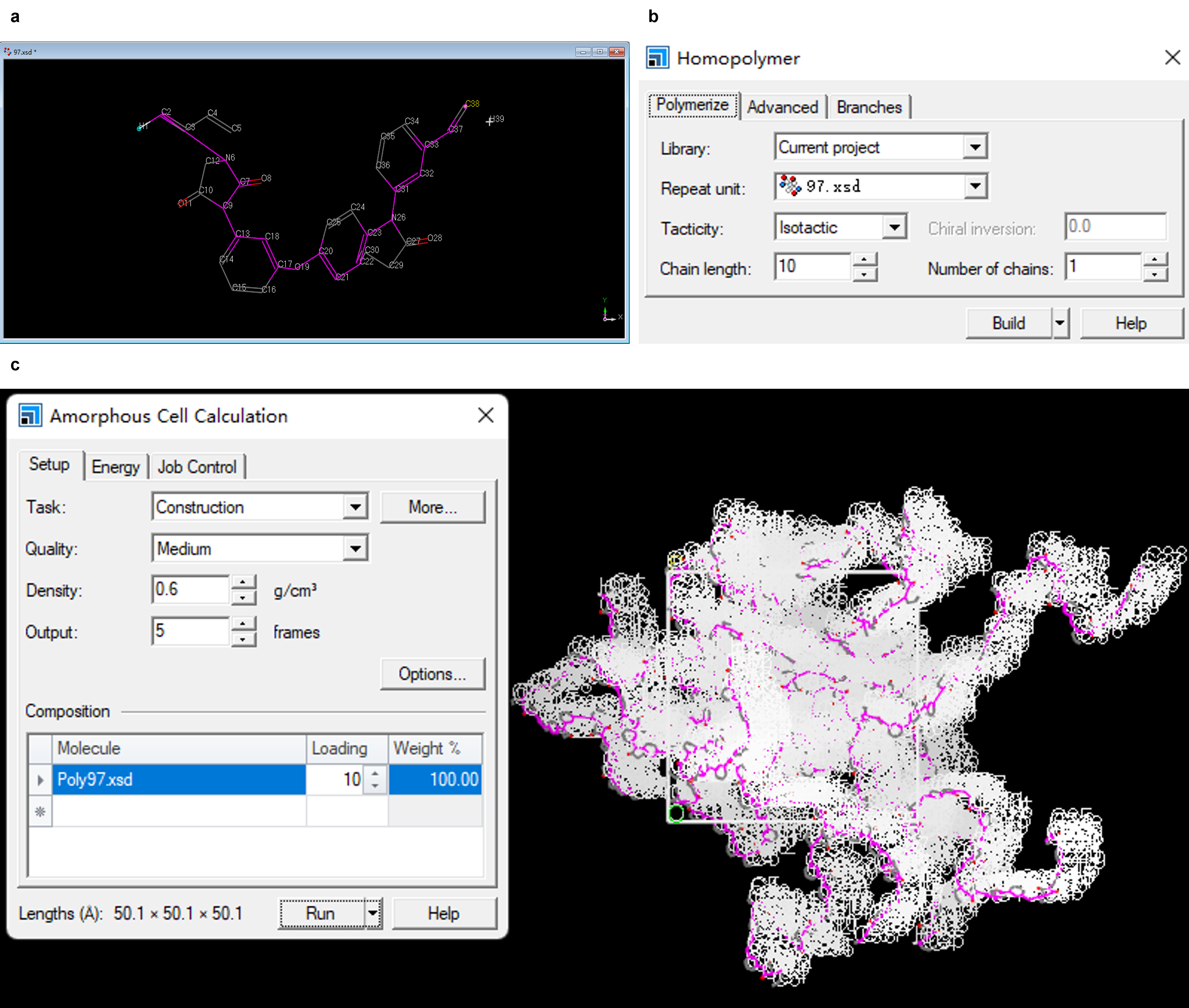}
    \caption{
    Construction of the polyimide amorphous periodic model.
    (a) Definition of the polyimide repeating unit and the first and
    last polymerization sites.
    (b) Construction of a linear homopolymer chain from the repeating
    units and packing of the chains into a three-dimensional periodic
    simulation box.
    (c) Amorphous polymer model generated with an initial density of
    \(0.6~\mathrm{g/cm^{-3}}\).
    }
    \label{fig:amorphous-construction}
\end{figure}

To reduce the influence of a single initial conformation on subsequent
simulation results, the main-chain dihedral angles were randomly sampled
during polymer-chain generation, allowing different chain segments to
adopt diverse initial conformations, as illustrated in
Figure~\ref{fig:amorphous-construction}(a)--(b).

After the polymer chains were constructed, they were randomly packed
into an amorphous simulation box with three-dimensional periodic
boundary conditions, as shown in
Figure~\ref{fig:amorphous-construction}(c). Each simulation system
contained 10 identical polymer chains with an initial density of
\(0.6~\mathrm{g/cm^{-3}}\).

The build quality of the amorphous model was set to medium, and five
candidate initial configurations were generated for each polyimide
system. The relatively low initial packing density reduced severe
atomic overlap during random packing and provided sufficient
conformational space for subsequent chain rearrangement, system
compression, and density equilibration.

The COMPASS force field was used during model construction, together
with the atomic charges and force-field types assigned to the current
structure.

\FloatBarrier

\subsubsection{Geometry Optimization and Cyclic Annealing}
\label{appendix:geometry-annealing}

\begin{figure}[!t]
    \centering
    \includegraphics[width=\columnwidth]{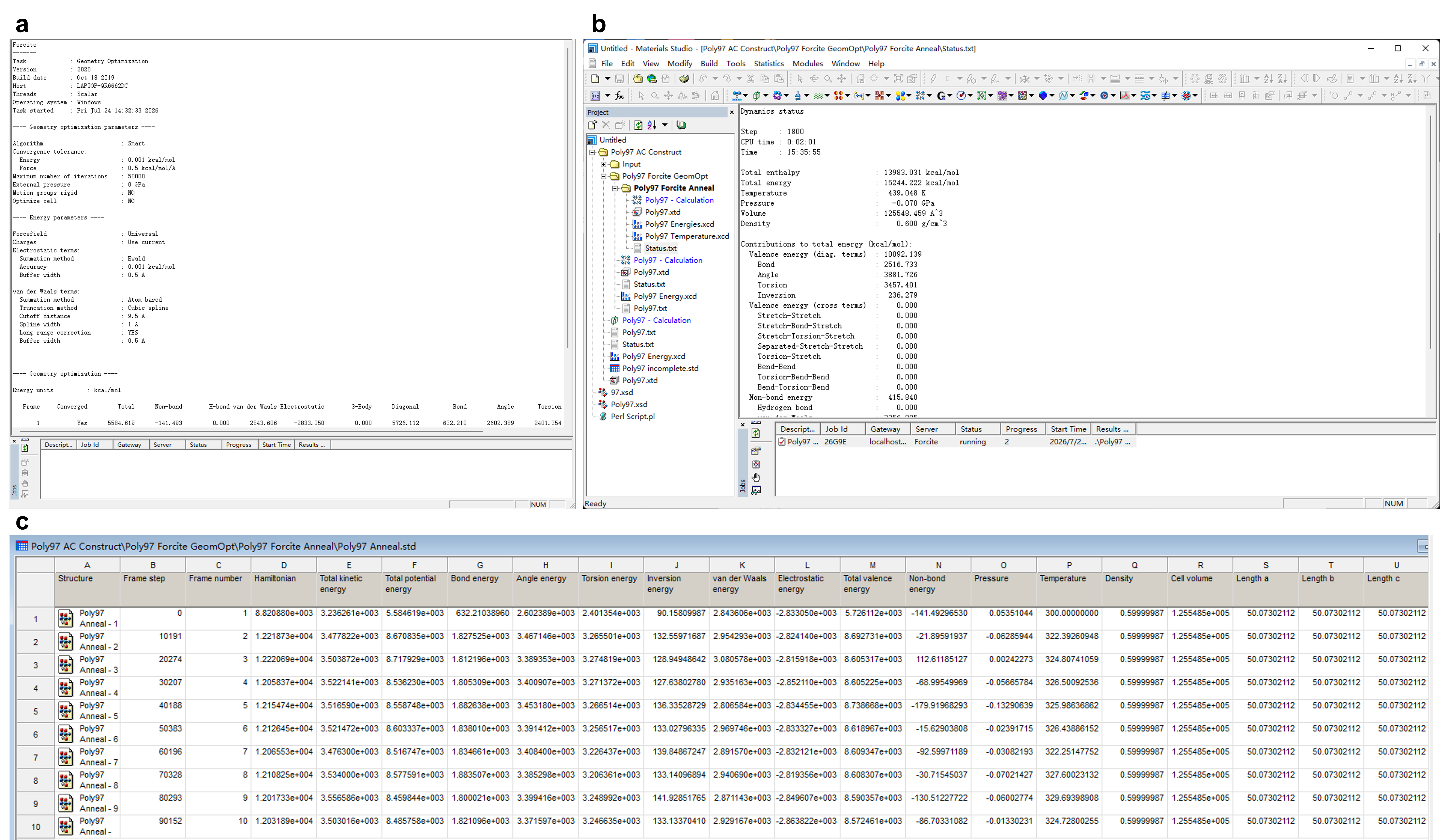}
    \caption{
    Structural optimization and annealing equilibration of the
    polyimide amorphous model.
    (a) Geometry optimization using the Smart algorithm.
    (b) Cyclic annealing performed over the temperature range of
    \(300\)--\(800~\mathrm{K}\).
    (c) Stable amorphous structure obtained after geometry optimization
    and cyclic annealing.
    }
    \label{fig:geometry-annealing}
\end{figure}

After the amorphous periodic model was constructed, geometry
optimization and cyclic annealing were performed, as illustrated in
Figure~\ref{fig:geometry-annealing}.

Geometry optimization was performed using the Smart algorithm in the
Forcite module. The maximum number of iterations was set to 50000,
and the van der Waals interaction cutoff radius was set to
\(9.5~\text{\AA}\).

During optimization, the atomic coordinates of the system were
continuously adjusted to reduce the total potential energy contributed
by bond stretching, bond-angle bending, dihedral torsion, and
non-bonded interactions. This procedure eliminated unreasonable atomic
contacts and local high-energy conformations generated during the
random packing of polymer chains, resulting in a more physically
reasonable initial amorphous structure.

After geometry optimization, the system was further subjected to
cyclic annealing. The annealing simulation was performed under the NVT
ensemble for 10 heating--cooling cycles. In each cycle, the system was
heated from \(300~\mathrm{K}\) to \(800~\mathrm{K}\) and subsequently
cooled to the initial temperature.

The heating process was divided into 25 temperature intervals, with
200 molecular dynamics steps performed at each interval. The
integration time step was set to \(1.0~\mathrm{fs}\).

\subsubsection{Multi-Temperature NVT and NPT Equilibration}
\label{appendix:nvt-npt-equilibration}

After geometry optimization and cyclic annealing, multi-temperature
molecular dynamics simulations were conducted within a preset
temperature range to obtain the equilibrium density of the polyimide
system and its temperature-dependent relationship~\citep{marti2024biopolymers,suter2025rapid}.

The simulation temperature range was centered on the estimated Tg of
the polymer and extended toward both higher and lower temperatures.
Multiple discrete temperature points were established at intervals of
\(20\,^{\circ}\mathrm{C}\).

At each temperature point, kinetic equilibration was first performed
under the NVT ensemble, allowing the system to reach a stable state
near the target temperature while maintaining a fixed simulation-cell
volume. The NVT simulation employed the COMPASS force field and a
Nosé thermostat. A total of 300000 molecular dynamics steps were
performed at each temperature, corresponding to a simulation time of
\(300~\mathrm{ps}\).

\begin{figure}[!htbp]
    \centering
    \includegraphics[
        width=\linewidth,
        height=0.76\textheight,
        keepaspectratio
    ]{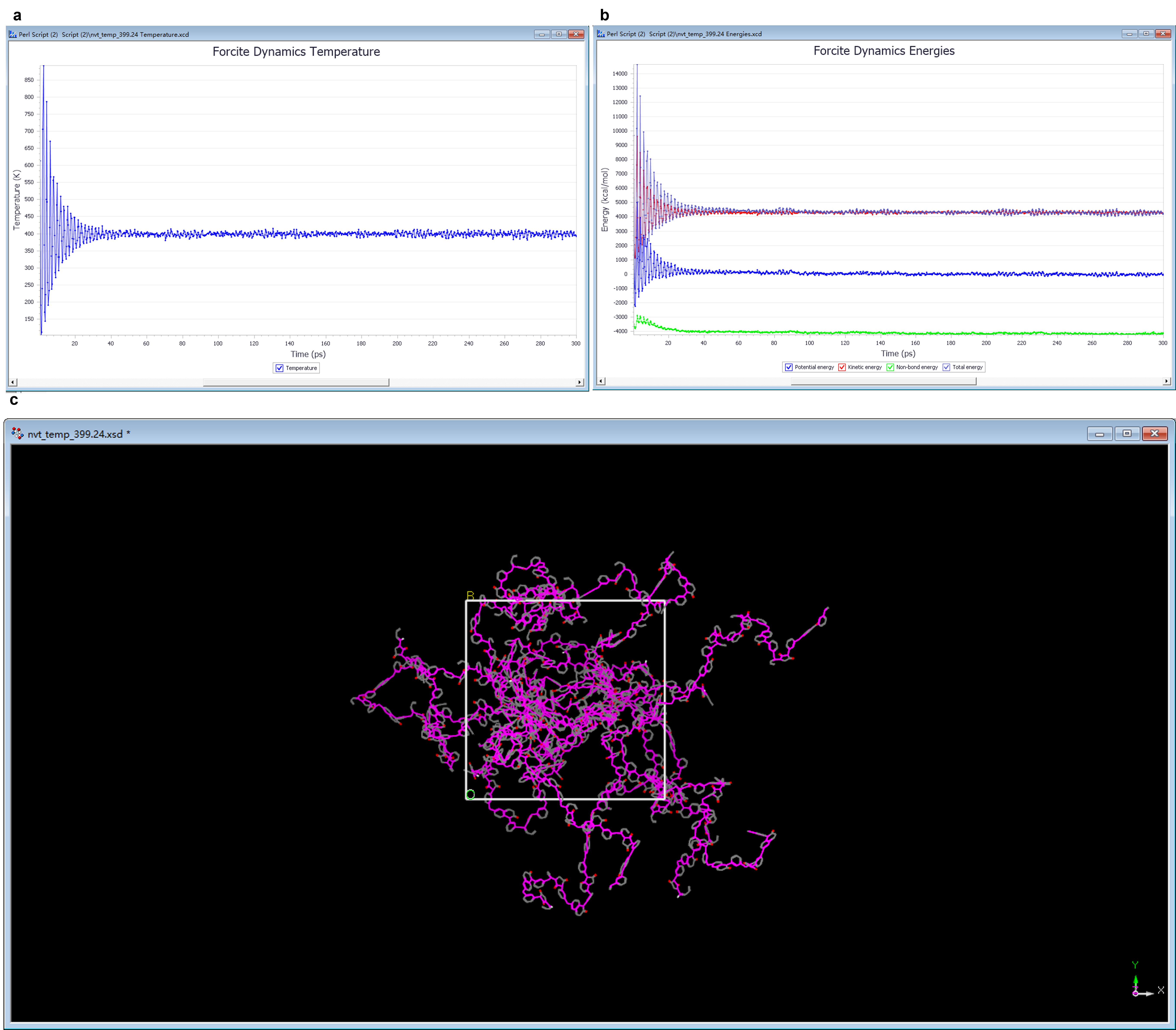}
    \caption{
    Representative NVT equilibration process before density extraction.
    (a) Temperature evolution during the NVT simulation.
    (b) Variations in potential, kinetic, non-bonded, and total energies.
    (c) Equilibrated amorphous periodic structure.
    }
    \label{fig:nvt-equilibration}
\end{figure}

During the NVT simulation, system equilibration was evaluated by
monitoring changes in temperature and energy components over time.
As shown in Figure~\ref{fig:nvt-equilibration}(a), the representative
system exhibited relatively large temperature fluctuations during the
initial stage of the simulation. These fluctuations were associated
with the initial velocity distribution and the rapid adjustment of
polymer-chain segments.

As the simulation progressed, the fluctuation amplitude gradually
decreased, and the system temperature eventually stabilized near the
target temperature. Correspondingly, the potential energy, kinetic
energy, non-bonded energy, and total energy gradually changed from
large initial fluctuations to stable fluctuations without persistent
drift, as shown in Figure~\ref{fig:nvt-equilibration}(b). These results
indicate that the system reached a stable thermodynamic state at the
current temperature.

The amorphous periodic structure obtained after NVT equilibration is
shown in Figure~\ref{fig:nvt-equilibration}(c). This structure was
subsequently used as the initial configuration for the NPT simulation
at the same temperature.

Following NVT equilibration, NPT molecular dynamics simulations were
performed at each temperature point. While maintaining a constant
target temperature, an external pressure of \(0.0001~\mathrm{GPa}\)
was applied to the system, allowing the volume of the periodic
simulation cell to adjust automatically according to the thermal
motion of the polymer chains.

Compared with fixed-volume NVT simulations, NPT simulations allow the
system to undergo thermal expansion or contraction and therefore
provide more physically reasonable equilibrium volumes and average
densities at different temperatures.

At high temperatures, polymer-chain motion is enhanced, the free volume
increases, and the system density is relatively low. As the temperature
decreases, chain motion becomes progressively restricted, interchain
packing becomes denser, and the system density increases accordingly.

For each temperature point, the average density was extracted only
from the stable portion of the NPT trajectory. The resulting average
densities were used to establish the density--temperature dataset
required for subsequent Tg fitting.

\FloatBarrier

\subsubsection{Determination of the Glass Transition Temperature Based on the Density--Temperature Relationship}
\label{appendix:tg-fitting}

To determine the simulated glass transition temperature, the average
density of the steady-state portion of the NPT trajectory was extracted
at each simulated temperature, producing the following
temperature--density dataset:

\begin{equation}
\label{eq:density-dataset}
\mathcal{D}_{\rho}
=
\left\{
\left(T_i,\rho_i\right)
\right\}_{i=1}^{N_T},
\end{equation}

where \(T_i\) represents the \(i\)-th simulated temperature,
\(\rho_i\) represents the average density obtained from the NPT
simulation at that temperature, and \(N_T\) represents the number of
temperature points used for fitting.

A continuous piecewise linear fit was subsequently performed using all
temperature--density data to describe the different thermal expansion
behaviors of the polymer in the low-temperature glassy state and the
high-temperature rubbery state~\citep{gudla2024determine,suter2025rapid}. The corresponding density--temperature
relationship is expressed as:

\begin{equation}
\label{eq:piecewise-density}
\rho(T)
=
\begin{cases}
a_{\mathrm{g}}T+b_{\mathrm{g}} 
& T\leq T_{\mathrm{g}}^{\mathrm{MD}}\\
a_{\mathrm{r}}T+b_{\mathrm{r}}
& T>T_{\mathrm{g}}^{\mathrm{MD}},
\end{cases}
\end{equation}

where \(a_{\mathrm{g}}\) and \(a_{\mathrm{r}}\) are the slopes of the
two fitted lines, while \(b_{\mathrm{g}}\) and \(b_{\mathrm{r}}\) are
the corresponding intercepts.

To ensure that the density--temperature relationship remains
continuous at the transition point, the two linear relationships
satisfy:

\begin{equation}
\label{eq:density-continuity}
a_{\mathrm{g}}T_{\mathrm{g}}^{\mathrm{MD}}
+b_{\mathrm{g}}
=
a_{\mathrm{r}}T_{\mathrm{g}}^{\mathrm{MD}}
+b_{\mathrm{r}}.
\end{equation}

Accordingly, the glass transition temperature obtained from the
molecular dynamics simulations is calculated as:

\begin{equation}
\label{eq:tg-intersection}
T_{\mathrm{g}}^{\mathrm{MD}}
=
\frac{
b_{\mathrm{r}}-b_{\mathrm{g}}
}{
a_{\mathrm{g}}-a_{\mathrm{r}}
}.
\end{equation}

During the fitting process, the squared error between the observed
densities at all temperature points and the densities predicted by the
piecewise linear model was minimized. The fitting objective is
expressed as:

\begin{equation}
\label{eq:tg-fitting-objective}
\widehat{T}_{\mathrm{g}}^{\mathrm{MD}}
=
\underset{T_{\mathrm{g}}}{\operatorname*{argmin}}
\sum_{i=1}^{N_T}
\left[
\rho_i-
\widehat{\rho}
\left(
T_i;T_{\mathrm{g}}
\right)
\right]^2.
\end{equation}

\begin{figure}[!t]
    \centering
    \includegraphics[width=0.5\columnwidth]{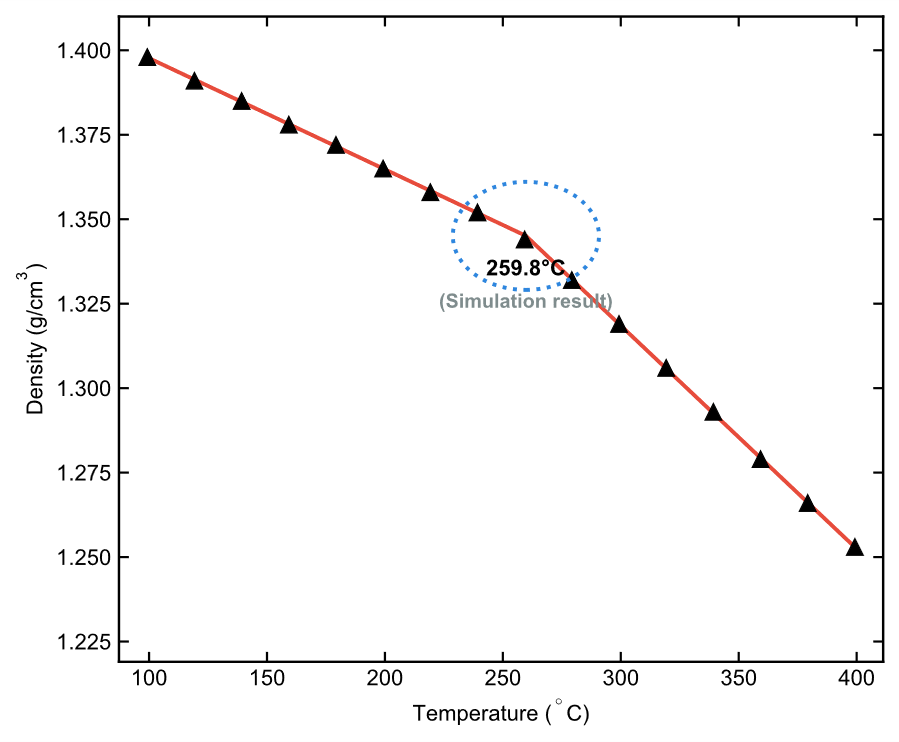}
    \caption{
    Determination of the simulated glass transition temperature by
    continuous piecewise linear fitting of the density--temperature
    relationship. The black triangles represent the average densities
    obtained from NPT simulations at different temperatures, and the
    red lines represent the fitted low-temperature glassy-state and
    high-temperature rubbery-state relationships.
    }
    \label{fig:tg-fitting}
\end{figure}

As shown in Figure~\ref{fig:tg-fitting}, the black triangles represent
the average densities obtained from NPT simulations at different
temperatures, while the red lines represent the continuous piecewise
linear fitting results.

As the temperature increases, the overall system density decreases.
However, the rate of density change varies significantly near
\(260\,^{\circ}\mathrm{C}\), reflecting the transition of the polymer
from the low-temperature glassy state to the high-temperature rubbery
state. The transition point obtained from the piecewise fitting is:

\begin{equation}
\label{eq:representative-tg}
T_{\mathrm{g}}^{\mathrm{MD}}
=
259.8\,^{\circ}\mathrm{C}.
\end{equation}

Therefore, \(259.8\,^{\circ}\mathrm{C}\) was determined as the
molecular dynamics simulation Tg of this representative polyimide
structure and was used as the simulation label of the corresponding
structure in the PITg-Curated dataset.

\FloatBarrier

\bibliographystyle{unsrtnat}
\bibliography{references,Su}

\end{document}